# A Driver Fatigue Recognition Algorithm Based on Spatio-Temporal Feature Sequence

Chen Zhang[1,2], Xiaobo Lu[*1,2], Zhiliang Huang[1,2]

1. School of Automation, Southeast University, Nanjing 210096, China;
2. Key Laboratory of Measurement and Control of Complex Systems of Engineering, Ministry of Education, Southeast University, Nanjing, 210096, China;
*Corresponding Author, E-mail: xblu2013@126.com

*Abstract*—**Researches show that fatigue driving is one of the important causes of road traffic accidents, so it is of great significance to study the driver fatigue recognition algorithm to improve road traffic safety. In recent years, with the development of deep learning, the field of pattern recognition has made great development. This paper designs a real-time fatigue state recognition algorithm based on spatio-temporal feature sequence, which can be mainly applied to the scene of fatigue driving recognition. The algorithm is divided into three task networks: face detection network, facial landmark detection and head pose estimation network, fatigue recognition network. Experiments show that the algorithm has the advantages of small volume, high speed and high accuracy.**

## I. Introduction

This paper designs a fast fatigue state recognition algorithm based on spatio-temporal feature sequence, which can be mainly applied to the scene of fatigue driving recognition. On the premise of ensuring the detection accuracy, this algorithm greatly reduces the size of network and detection time, so it can be transplanted to the embedded platform with low computing power to meet the application requirements. The algorithm is mainly divided into three parts: face detection, facial landmark detection and head pose estimation, fatigue recognition.

First of all, fatigue recognition needs to detect the face from the image and mark it with bounding box, otherwise it is too difficult to get the facial landmark and head pose directly from the large scene. Face detection has always been a complex and challenging detection problem which needs to solve the multi-scale problem, robustness requirement, and the efficiency of face detection. The existing face detection and alignment algorithm with high speed and high accuracy is MTCNN [1] proposed by Zhang. However, MTCNN still cannot meet the real-time requirements of fatigue recognition based on embedded platform, so corresponding improvement is needed.

Facial landmark detection and head pose estimation are very important for fatigue recognition. The wrong facial landmarks or head poses will seriously affect fatigue recognition accuracy. In recent years, deep learning has been used in facial landmark detection, and its effect is far better than traditional methods: Sun [2] proposed a CNN based on cascade structure for facial landmark detection. This method has good detection effect on various pose changes, expression changes and occlusion. Miao used Doubly CNN [3] to obtain facial landmark, its accuracy is higher than TCDCN but its efficiency is not satisfactory.

Recent studies have proved that the fatigue recognition method based on continuous multi-frame images is superior to the traditional method. The LRCN [4] proposed by Donahue provides a framework that can process single frame image and video stream simultaneously, which uses LSTM [5] to find the correlation information between continuous frames, so as to classify spatio-temporal feature vectors.

Based on the previous researches, this paper combines the advantages of each method. For applying on embedded platform, a real-time fatigue state recognition algorithm is designed. This paper is distinguished by the following main contributions:

- A face detection and alignment algorithm based on MTCNN is designed, which improves the detection speed and reduces the size of the model while ensuring the detection accuracy.

- Influenced by the idea of multi-task learning [6], a lightweight multi-task learning network based on minimum feature extraction unit is designed for facial landmark detection and head pose estimation.

- Inspired by LRCN, a LSTM-based network based on spatio-temporal fatigue feature sequence is designed for fatigue recognition.

The rest of the paper is organized as follows: Section II introduces our methods in detail. Section III presents the experimental settings and performance analysis. Finally, Section IV summarizes this paper.

## II. Method

### A. Face Detection and Alignment Network

First, fatigue recognition requires to detect and mark the face with bounding box from the image. MTCNN [1] proposed by Zhang is a multi-task learning network, which can process these two tasks simultaneously and be divided into three stages:

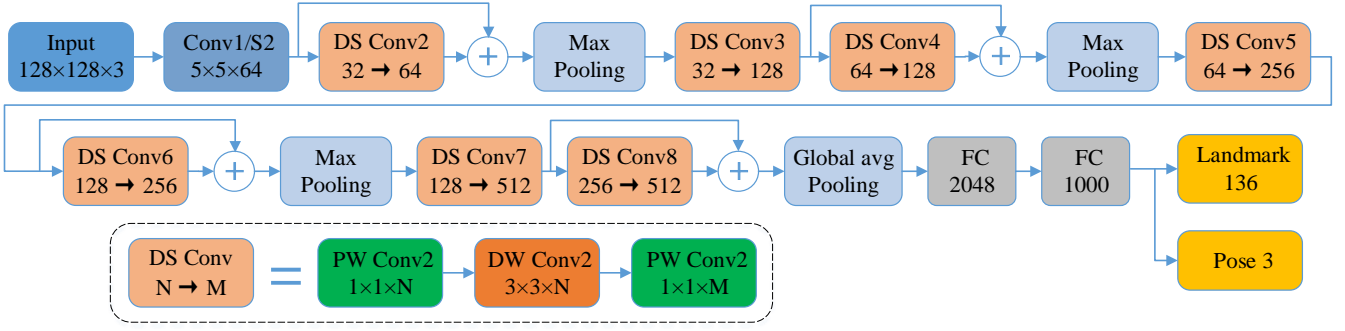

Figure 1. Facial landmark detection and head pose estimation network.
Chart in dotted box is the minimum feature extraction unit.

Proposal Network (P-Net), Refine Network (R-Net), Output Network (O-Net). MTCNN is efficient and fast, and shows strong robustness in engineering applications. However, its processing speed on embedded platforms is still not ideal, especially when it needs to be combined with other tasks.

Experimental results show that R-Net and O-Net, including fully connect layer, account for more than 98% of MTCNN computing time and most parameters. In this paper, the Global Average Pooling layer (GAP) [7] is adopted to replace the fully connect layer in R-Net and O-Net as the output layer. The GAP can reduce a large number of parameters and 50% forward computing time caused by the fully connect layer [8]. Our experiment will prove the effectiveness of this improvement.

### B. Facial Landmark Detection and Head Pose Estimation Network

This network is a multi-task learning network that uses the face detected by the previous face detection network to detect facial landmark and estimate head pose. In addition, a minimum feature extraction unit is designed to reduce network volume and detection time.

*1) Minimum Feature Extraction Unit:* The Depthwise Separable Convolution (DSC) [9] structure is adopted in this unit to decompose Standard Convolution (SC) operation into two operations: Depthwise Convolution (DWC) and Pointwise Convolution (PWC). Each convolution kernel in DWC only convolved one channel in the input image, instead of performing the same convolution operation on all channels like SC, the size of convolution kernel used by PWC is $1 \times 1 \times M$, where $M$ is the number of feature maps after DWC.

In addition, in order to avoid the disappearance of gradient in the deep network model, the unit adopts Shortcut Structure to connect the input and output. This structure allows the output to combine the original feature with the convoluted feature, which is equivalent to splicing the feature map onto the convolution layer. Furthermore, this unit uses LeakyReLU to replace ReLU for enhancing the fitting ability when using CNN to detect landmark, and the LeakyReLU parameter is 0.1.

In summary, a minimum feature extraction unit based on the above ideas is designed, which framework is shown in Figure 1.

*2) Multi-task Learning Network:* In single-task learning, the back propagation of gradient tends to fall into local minimum value, which is easy to cause network overfitting. However, multi-task learning can use the implicit correlation between tasks to improve the generalization ability of the network. Influenced by this idea, we assume that facial landmark detection and head pose estimation are relevant. Common facial landmark detection is divided into 5 points, 25 points, 68 points and 106 points. Considering our requirements and existing datasets, 68 points were selected. Head pose estimation refers to obtain head position by camera, which is defined by three degrees of freedom: Pitch Angle, Yaw Angle and Roll Angle.

*3) Loss Function:* The coordinate of each facial landmark is represented by two values, so the dimension of the facial landmark output layer is 136. The learning object is formulated as a nonlinear regression problem. For each landmark $y_i^{landmark}$, we use square loss function:

$$L_i^{landmark} = \sum_{i=1}^{136} \|\hat{y}_i^{landmark} - y_i^{landmark}\|_2^2 \quad (1)$$

Where $\hat{y}_i^{landmark}$ is the ground-truth of the facial landmark, $y_i^{landmark}$ is the output value of the network.

The head pose contains the rotation angle (Yaw, Roll, Pitch) of three axis, so the dimension of the output layer of the head pose estimation is 3 and its angle range is $(-90°, 90°)$. Similar with facial landmark detection, we use square loss function:

$$L_i^{pose} = \sum_{i=1}^{3} \|\hat{y}_i^{pose} - y_i^{pose}\|_2^2 \quad (2)$$

Where $\hat{y}_i^{pose}$ is the true value of the head pose angle, $y_i^{pose}$ is the output value of the network.

The total loss function:

$$L = \frac{1}{2}\alpha L_i^{landmark} + \frac{1}{2}\beta L_i^{pose} \quad (3)$$

Where $\alpha$ and $\beta$ are the weight of the loss functions. In our experiments, both $\alpha$ and $\beta$ are 0.5.

In summary, this multi-task network uses the minimum feature extraction unit to detect facial landmarks and estimate head pose. Although the network has a multi-layer structure, the unit greatly reduces the number of parameters in the model. The parameter value of the whole network model is calculated only 3723896, which is about 14.2MB. Its framework is shown in Figure 1, and the test result is shown in Figure 2.

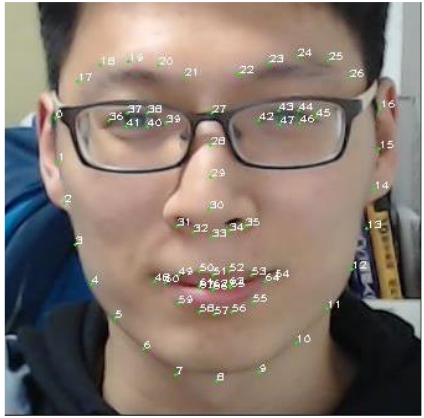

Figure 2. Facial landmark detection result

*C. Spatio-Temporal Fatigue Feature Extraction*

In this paper, spatio-temporal fatigue feature sequence is designed for fatigue recognition. In order to construct the spatio-temporal fatigue feature sequence, the fatigue feature description of each face image should be obtained first. The spatio-temporal fatigue feature sequence designed in this paper utilizes eye, mouth and head pose features related to fatigue.

*1) Eye Fatigue Feature:* Research from Carnegie Mellon University shows that when people are fatigue, their blinking rate will increase and their eyes will stay closed for longer. In addition, PERCLOS index [10] is proposed to measure the fatigue degree of drivers, which is defined as the time occupied by the proportion of eye closure per unit of time. According to the research, we divide the eyes into two states: open and closed.

From Figure 2, it can be seen that the contour of the left eye is marked by 6 points from 36 to 41, while the contour of the right eye is marked by 6 points from 42 to 47. In order to avoid the incomplete human eyes caused by head tilt, the area of human eyes is increased to ensure the complete eye contour.

*2) Mouth Fatigue Feature:* Yawn is one of the important features to judge driver fatigue driving. When a person yawns, his mouth opens wide and stays open for a while. Therefore, the opening degree of the mouth is a direct indicator of the state of the mouth, so this paper selected the opening degree of the mouth as the feature of mouth fatigue.

Given that the actual thickness of the lips varies from person to person, the inner contour of mouth is more indicative of the opening degree of mouth than the outer contour. Therefore, the opening degree of the mouth can be defined as follows: The ratio of the maximum height to width of the inner contour of the mouth when the mouth is open.

*3) Head Pose Fatigue Feature:* When the driver is in the fatigue state, his head pose often has corresponding states, usually manifested as frequent nodding. Therefore, frequent nodding can be regarded as an important feature of fatigue.

Frequent nodding of the head when fatigue can be regarded as a continuous swing of the head around the Pitch axis, so the angle on the Pitch axis will have obvious fluctuations. The head pose estimation algorithm designed in this paper is used to process video that nods frequently when fatigue. Our experimental results show that, when people nod frequently due to fatigue, their head pose will have obvious fluctuations, which is mainly reflected in the change of Pitch axis angle. Therefore, this paper uses the angle of the Pitch axis as the description of driver's head pose fatigue feature.

*D. Fatigue recognition Network*

Traditional fatigue recognition methods usually use single fatigue feature. However, fatigue is continuous spatio-temporal actions, such as frequent blinking, yawning and frequent nodding. Therefore, simply analyzing the expression of a single facial image can easily lead to misjudgment. Inspired by the video behavior analysis structure LRCN [4], a two-stage fatigue recognition method is designed. Stage One: extract the fatigue features of each frame from video, and then splice these features to construct the spatio-temporal fatigue feature sequence. Stage Two: Recognize fatigue state by LSTM-based network based on spatio-temporal fatigue feature sequence.

*1) Construction of Multi-Feature Spatio-Temporal Fatigue Feature Sequence:* By extracting the facial fatigue feature of each frame from video, the fatigue feature vector with the length of 4 for each frame can be obtained. This vector contains the state of the left and right eyes, the opening degree of the mouth, and the angle of the head pose on the Pitch axis. The fatigue feature vector is shown in equation (4):

$$v_t = \left\{x_{l_{eye(t)}}, x_{r_{eye(t)}}, x_{mouth(t)}, x_{pose(t)}\right\}, t = \{0,1,2,\dots,N\} \quad (4)$$

Where $x_{l_{eye(t)}}$ and $x_{r_{eye(t)}}$ are the state of the left and right eyes, $x_{mouth(t)}$ is the opening degree of the mouth, $x_{pose(t)}$ is the angle of the head pose on the Pitch axis.

The spatio-temporal fatigue feature sequence is composed of fatigue feature vectors of continuous multi-frame images. A four-dimensional matrix can be constructed by splicing the fatigue feature vectors of each frame. The matrix dimension is the length of the fatigue feature vector in each frame, and the length is the number of frames included. The multi-frame image fatigue feature vector splicing process is shown in Figure 3:

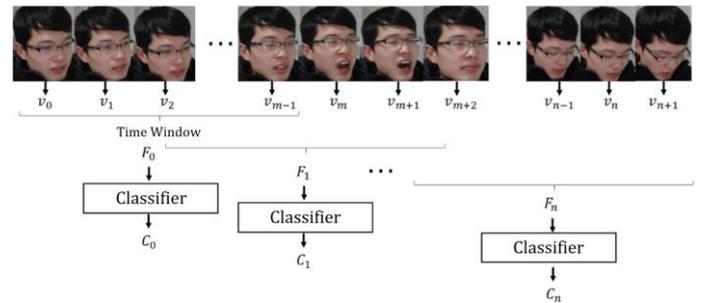

Figure 3. Construction of spatio-temporal fatigue feature sequence

As can be seen from Figure 3, the segments in the video can be encoded into independent spatio-temporal fatigue feature sequences by using the time window. Assuming that the length of the sliding window is $N$, a matrix of $4 \times N$ will be obtained, and the height of the matrix is the size of the time window, that is, the length of the obtained sequence, and the width is the dimension of the facial fatigue feature of each frame. In order to reduce the redundancy calculation, the spatio-temporal fatigue feature sequences are stored in double-end queues. When extracting the fatigue feature vector of the new frame image, the fatigue feature vector at the front of the queue will pop up, and then the new fatigue feature vector will be put into the end of the queue. This process will repeat until the end, and its calculation formulations are as follow:

$$C_t = \{v_{t_0}, v_{t_0+1}, v_{t_0+2}, \dots, v_{t_0+n}\} \quad (5)$$

$$Deque(C_t) = \{v_{t_0+1}, v_{t_0+2}, \dots, v_{t_0+n}\} \quad (6)$$

$$Enque(C_t) = \{v_{t_0+1}, v_{t_0+2}, \dots, v_{t_0+n+1}\} \quad (7)$$

The length of time window is the key parameter to construct the time-space fatigue characteristic sequence. If the length is too short, the obtained spatio-temporal fatigue feature sequence may not fully cover the fatigue action, while the overlong time window will result in the sequence containing too much redundant information, bringing excessive calculations. Therefore, the selection of the length of time window will affect the accuracy and efficiency of the fatigue recognition algorithm.

The number of skipped frames is another key parameter. Skipping frame detection can reduce the length of the spatio-temporal fatigue feature sequence, reduce the complexity of system, and improve the real-time performance of the system. The spatio-temporal fatigue feature sequence with skipping frame detection is shown in equation (8):

$$F_i = \{v_t, v_{t+k}, v_{t+2k}, \dots, v_{t+nk}\} \quad (8)$$

Where $n$ is the time window length, $k$ is the number of skipped frames. In this paper, these two optimal parameters will be obtained through contrast experiment.

*2) Fatigue Recognition Network:* Compared with RNN, LSTM-based network uses cumulative form to compute cell, thus avoiding the gradient disappearance problem. In addition, it has better performance in sequence data classification and recognition task because of its long-term memory. Based on the above ideas, this paper designs a fatigue recognition network based on LSTM, and its structure is shown in Figure 4:

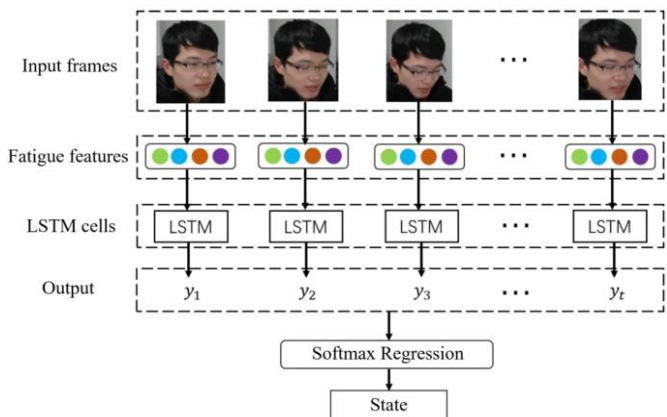

Figure 4. Fatigue Recognition Network

Firstly, this network extracts the spatio-temporal fatigue feature from single frame in the video. Then, construct the spatio-temporal fatigue feature sequence by splicing these feature vectors. Finally, use the fatigue recognition network based on LSTM to recognize these sequences. Thus, fatigue state recognition is transformed into a spatio-temporal sequence recognition problem.

*3) Loss Function:* In this paper, fatigue recognition is a binary classification problem, we use the cross-entropy loss:

$$L_i^{cls} = -\left(y_i^{cls} \log(pi) = (1 - y_i^{cls})(1 - \log(p_i))\right) \quad (9)$$

Where $p_i$ is the probability that this sample is fatigue state, $y_i^{cls}$ is the ground-truth.

III. EXPERIMENT

A. Evaluation of Face Detection Network

The datasets, loss functions and experimental methods used in this paper are similar with MTCNN [1], except for using the GAP to replace the fully connect layer.

*1) Accuracy of Face Detection:* This paper uses the published FDDB [11] evaluation set to detection accuracy, and uses discrete counting method to measure the algorithm accuracy. Experiment results shows that the accuracy of our method is 92.1%, which reaches the mainstream detection accuracy level.

*2) Speed of Face Detection:* This experiment uses Caffe to compare the detection time of each frame of this algorithm with the mainstream face detection algorithms on the NVIDIA Jetson TX2 embedded platform. The results are shown in TABLE I.

TABLE I. FACE DETECTION SPEED

| Algorithms | Average Detection Time(s) | Frame(fps) |
|---|---|---|
| Adaboost [12] | 0.378 | 2.6 |
| Faster R-CNN [11] | 0.185 | 5.4 |
| UnitBox [13] | 0.233 | 4.3 |
| Cascade-CNN [14] | 0.061 | 16.4 |
| MTCNN [1] | 0.046 | 21.7 |
| **Our Method** | **0.031** | **32.3** |

The results prove that our method improved the detection speed and reduced the volume of the model on the premise of ensuring the detection accuracy.

B. Evaluation of Facial Landmark Detection and Head Pose Estimation Network

*1) Datasets:* The experiment uses the 300-W [15] dataset as the training set. This paper picks out IBUG, LFPW and HELEN test set in 300-W as the total test set. The results of IBUG are called challenge set results, the result of LFPW and HELEN are called ordinary set results.

*2) Facial Landmark Detection:* In this paper, average error based on binocular distance normalization is adopted to measure the accuracy of facial landmark detection algorithm. The formula is as follows:

$$err = \frac{1}{N}\sum_{i=1}^{N}\frac{\frac{1}{M}\Sigma_j^M |p_{i,j} - y_{i,j}|_2}{|l_{i,j} - r_{i,j}|_2} \quad (10)$$

Where $N$ is the number of test samples, $M$ is the number of landmarks, $p_{i,j}$ is the facial landmarks predicted by network, $y_{i,j}$ is the ground-truth landmarks, $l_{i,j}$ is the landmark of left eye corner, and $r_{i,j}$ is the landmark of the right eye corner.

This algorithm is compared with the mainstream 68-point facial landmark detection algorithm. In order to ensure the correlation between facial landmark detection and head pose estimation, we designed the corresponding experiment, and the results are shown in TABLE II.

TABLE II. LANDMARK DETECTION ACCURACY

| Model | Ordinary set(%) | Challenge set(%) | Total set(%) |
|---|---|---|---|
| CFAN [16] | 5.50 | 16.78 | 7.69 |
| SDM [17] | 5.60 | 15.40 | 7.52 |
| ESR [18] | 5.28 | 17.00 | 7.58 |
| LBF [19] | **4.95** | 11.98 | 6.32 |
| Our model (except head pose estimation) | 5.02 | 11.25 | 6.24 |
| **Our complete model** | 4.98 | **10.72** | **6.10** |

As can be seen from TABLE II, the average detection error of our method is smaller than that of other algorithms, especially in challenge set. It proves that our method is more suitable for scenarios with large pose changes. In addition, we prove the correlation between facial landmark detection and head pose estimation by comparing the results of the last two rows.

*3) Head Pose Estimation:* In order to verify the performance of the network, we use the average absolute error as the criterion to measure the accuracy of head pose estimation, and the formula is as follows:

$$MAE = \frac{1}{N}\sum_{i=1}^{N}|f_i - y_i| \quad (11)$$

Where $N$ is the number of test samples, $f_i$ is the head pose predicted by network, and $y_i$ is the ground-truth head pose.

We compared our algorithm with the head pose estimation network proposed by Yang [20] on the 689 test sets in 300-W. The results are shown in TABLE III.

TABLE III. POSE ESTIMATION ACCURACY

| Method | Pitch (°) | Yaw (°) | Roll (°) |
|---|---|---|---|
| Yang [20] | 5.199 | 4.203 | 2.426 |
| **Our Method** | **3.812** | **2.374** | **1.378** |

It can be seen from TABLE III that the average error of our method in 3 dimensions are lower than that of Yang's algorithm. The results show that multi-task learning can improve the accuracy of head pose estimation to some extent.

## C. Evaluation of Fatigue Recognition Network

*1) Dataset:* This paper uses the fatigue video shoot by our laboratory as training and test samples. The samples contain 10 video segments, and each video segment contains fatigue and normal states. Fatigue states include yawning, long periods of eye closing, frequent nodding, and the videos also contain distracting behaviors such as talking and smiling. Fatigue states and normal states are contained in the same video segment. Corresponding video frames is marked manually, and then use time windows to generate training samples. In this paper, the overlapping frame number is used to determine whether the video segment is in fatigue state. If the number of fatigue state frames are accounted for more than 80% of the total video frames in this segment, this video segment will be labeled as fatigue video sample; otherwise, this video segment will be labeled as normal video sample.

According to the statistical analysis of video samples, a fatigue action lasts about 3-5 seconds. We set the length of each sample to 5 seconds to ensure complete coverage of fatigue effects. Since the video frame rate in this paper is 30 frames per second, the total number of frames for each video segment is 150 frames. In this paper, all the videos are processed by sliding window, and a total of 4,298 training and test video segment samples are obtained. There are 1,062 samples classified as fatigue state and 2,680 samples classified as wakeful state. The ratio of positive and negative samples is about 1:3.

*2) Fatigue Recognition Network:* In this paper, fatigue recognition network shown in Figure 4 is used to explore the optimal network parameters: the length of sliding window $N$ and the number of skipped frames $k$.

Different length of time window contains different amount of fatigue feature information. The longer the time window is, the more fatigue information it contains, but also needs more additional computation. In this paper, spatio-temporal fatigue feature sequences with the lengths of time window of 60, 90, 120 and 150 are selected for network training and testing to test fatigue recognition accuracy and average detection time. This experiment is run on the NVIDIA Jetson TX2 embedded platform, and results are shown in TABLE IV:

TABLE IV. SLIDE WINDOW LENGTH EXPERIMENT

| Plan | Length (frames) | Accuracy (%) | Time (ms) |
|---|---|---|---|
| A | 150 | 91.1 | 128.7 |
| B | 120 | 91.6 | 105.1 |
| C | 90 | 88.3 | 77.44 |
| D | 60 | 86.2 | 53.81 |

Skipping frame detection can reduce the length of the spatio-temporal fatigue feature sequence, reduce the complexity of system, and improve the real-time performance of the system. In order to determine the effect of the number of skipped frames on the detection accuracy and average detection time, this paper selects 120 frames as the length of sliding window as baseline, and selects the skipped intervals of 0,1,2,3,4 respectively for comparison. This experiment is run on the NVIDIA Jetson TX2 embedded platform, and results are shown in TABLE V:

TABLE V. SKPPED FRAME INTERVAL EXPERIMENT

| Plan | Intervals (frames) | Length (frames) | Accuracy (%) | Time (ms) |
|---|---|---|---|---|
| A | 0 | 120 | 91.6 | 105.1 |
| B | 1 | 60 | 91.0 | 52.97 |
| C | 2 | 40 | 88.3 | 38.34 |
| D | 3 | 30 | 86.2 | 31.81 |
| E | 4 | 20 | 80.2 | 21.10 |

Experiment results show that shortening the length of time window and skipping frame can reduce the detection time. However, the detection accuracy of the recognition network will decline if skip too many frames or shorten too much length of time window. Because the fatigue feature sequence contains too less feature information to recognize the fatigue state.

Based on the above experimental results, the network with the length of time window of 60 frames and the skipped interval of 1 frame has better detection accuracy and lower average detection time. Finally, Plan B in TABLE V is the most suitable plan for Jetson TX2.

IV. CONCLUSION

This paper designs a real-time fatigue state recognition algorithm based on spatio-temporal feature sequence, which can be mainly applied to the scene of fatigue driving recognition. The algorithm is divided into three task networks: face detection network, facial landmark detection and head pose estimation network, fatigue recognition network. In addition, corresponding experiments are designed to explore two optimal key parameters: the length of sliding window and the number of skipped frames.

In this paper, the networks are cascaded and transplanted to embedded platform NVIDIA Jetson TX2 because of its small volume, high speed and high accuracy.

ACKNOWLEDGMENT

This work was supported by the National Natural Science Foundation of China (No.61871123), Key Research and Development Program in Jiangsu Province (No.BE2016739) and a Project Funded by the Priority Academic Program Development of Jiangsu Higher Education Institutions.